\documentclass[conference,compsoc]{IEEEtran}
\usepackage{blindtext, graphicx}

\ifCLASSINFOpdf

\usepackage[tight,footnotesize]{subfigure}
\usepackage{textcomp}
\usepackage{float}
\usepackage{graphicx}
\usepackage{epstopdf}
\usepackage{color}
\usepackage{wrapfig,lipsum}
\usepackage{setspace}
\usepackage{graphics}
\usepackage{grffile}
\usepackage{lipsum,multicol}
\usepackage{tabularx}
\usepackage{array}
\usepackage{algorithmic}
\usepackage{algorithm}
\usepackage{lineno}
\usepackage{mathptmx}
\usepackage{latexsym}
\usepackage{moreverb}
\usepackage{amsmath}
\usepackage{calc}
\usepackage{amssymb}
\usepackage{sidecap}
\usepackage{verbatim}
\usepackage{longtable}
\usepackage{multirow}
\usepackage{color,soul}
\usepackage{multirow}
\usepackage{microtype}
\usepackage{enumitem}

\definecolor{darkred}{RGB}{200, 50, 50}

\usepackage[style=ieee,minbibnames=1,maxbibnames=2,doi=false,isbn=false,url=false,eprint=false,date=year,bibencoding=utf8]{biblatex}
\AtEveryBibitem{\clearfield{pages}}
\AtEveryBibitem{\clearfield{volume}}
\AtEveryBibitem{\clearfield{number}}
\AtEveryBibitem{\clearfield{note}}
\AtEveryBibitem{\clearlist{language}}

\IEEEoverridecommandlockouts
\newcolumntype{C}[1]{>{\centering\let\newline\\\arraybackslash\hspace{0pt}}m{#1}}
\newcommand{\cmmnt}[1]{\ignorespaces}
\begin{document}
	
	%
	\title{
	\vspace{-0.6cm}
	{\fontsize{18.6}{23}\selectfont Combining General and Personalized Models \\ for Epilepsy Detection with Hyperdimensional Computing}
	\vspace{-0.8cm}
	
	\thanks{This work has been partially supported by the ML-Edge Swiss National Science Foundation (NSF) Research project (GA No. 200020182009/1), and the PEDESITE Swiss NSF Sinergia project (GA No. SCRSII5 193813/1). T. Teijeiro is supported by the grant RYC2021-032853-I funded by MCIN/AEI/ 10.13039/501100011033 and by the "European Union NextGenerationEU/PRTR"}
	}
	
	
		
	
	\author{
	\IEEEauthorblockN{Una Pale\IEEEauthorrefmark{1}, Tomas Teijeiro\IEEEauthorrefmark{1}\IEEEauthorrefmark{2}, David Atienza\IEEEauthorrefmark{1}}
	\IEEEauthorblockA{\IEEEauthorrefmark{1}Embedded Systems Laboratory (ESL), Ecole Polytechnique Federale de Lausanne (EPFL), Switzerland\\ \IEEEauthorrefmark{2} BCAM - Basque Center for Applied Mathematics, Spain \\ 
		\{una.pale, david.atienza\}@epfl.ch, tteijeiro@bcamath.org}
	
		\vspace{-0.8cm}
		}

	\maketitle
	
    \begin{abstract}

    Epilepsy is a chronic neurological disorder with a significant prevalence. However, there is still no adequate technological support to enable epilepsy detection and continuous outpatient monitoring in everyday life. Hyperdimensional (HD) computing is an interesting alternative for wearable devices, characterized by a much simpler learning process and also lower memory requirements. 
    In this work, we demonstrate a few additional aspects in which HD computing, and the way its models are built and stored, can be used for further understanding, comparing, and creating more advanced machine learning models for epilepsy detection. These possibilities are not feasible with other state-of-the-art models, such as random forests or neural networks. 
    We compare inter-subject similarity of models per different classes (seizure and non-seizure), then study the process of creation of generalized models from personalized ones, and in the end, how to combine personalized and generalized models to create hybrid models. This results in improved epilepsy detection performance. We also tested knowledge transfer between models created on two different datasets. Finally, all those examples could be highly interesting not only from an engineering perspective to create better models for wearables, but also from a neurological perspective to better understand individual epilepsy patterns. 
		
	\end{abstract}
	
	\IEEEpeerreviewmaketitle

	\section{Introduction}
	\label{sec:intro}
    \vspace{-2mm}
    Epilepsy is a chronic neurological disorder characterized by the unexpected occurrence of seizures, imposing serious health risks and many restrictions on the daily life of patients. It affects a significant portion of the world's population (0.6 to 0.8\%)~\cite{mormann_seizure_2007}, of which one third still suffer from seizures despite pharmacological treatments. Thus, a potential solution is to have small, non-stigmatizing, wearable devices for long-term epilepsy monitoring in patients' homes and everyday life rather than limited to in-hospital monitoring.    
    
    Although many studies report impressive levels of accuracy in epilepsy detection using machine learning (ML) methods, widespread adoption of commercial technology has not yet occurred. The reasons for this are many, some of which are specific to the complexity of epilepsy itself, such as spontaneous occurrence, unbalanced datasets, and multimodal nature.
    Furthermore, seizures show highly personalized patterns, which require new methods of personalizing general models using the characteristics of individual patients~\cite{sopic_personalized_2022}.
    Other aspects that need to be taken into account when designing algorithms for future lightweight wearable devices with long battery life are computational complexity and memory consumption. Many state-of-the-art ML algorithms for epilepsy detection are, for this reason, not ideal for wearable devices. Thus, hyperdimensional computing is posed as a potential alternative. 

    Hyperdimensional computing (HDC) is an emerging ML paradigm inspired by neuroscience research, based on data representation in the shape of high-dimensional \textit{hypervectors} (usually having more than 10000 dimensions)~\cite{kanerva_hyperdimensional_2009}. 
    The paradigm shift in data representation brings various advantages for efficient learning and low-power hardware implementation. From a learning perspective, it opens new paths for semi-supervised~\cite{imani_semihd_2019}, distributed~\cite{imani_framework_2019}, continuous online learning~\cite{benatti_online_2019}, or multicentroid learning~\cite{pale_multi-centroid_2022}. In terms of hardware, parallelization possibilities open the way to design efficient accelerators~\cite{imani_revisiting_2021} or in-memory computations~\cite{karunaratne_energy_2021}. Its lower energy and memory requirements enable learning in low-power wearables and IoT systems. 
    HD computing has attracted a great deal of attention for various biomedical applications, one of them being epilepsy detection~\cite{burrello_ensemble_2021,asgarinejad_detection_2020, pale_exploration_2022}. 
    Most of the HDC models developed by researchers are either personalized (only using patient-specific data) or generalized (using all patients) and are rarely discussed or compared. Thus, in this work we want to demonstrate several interesting aspects of HD computing models, such as the ability to create generalized models from personalized ones, compare them, and also to build hybrid models from personalized and generalized ones at the same time.  

	\begin{figure*}
		\centering
		\includegraphics[trim={0cm 0cm 0cm 0cm},clip,width=0.95\linewidth]{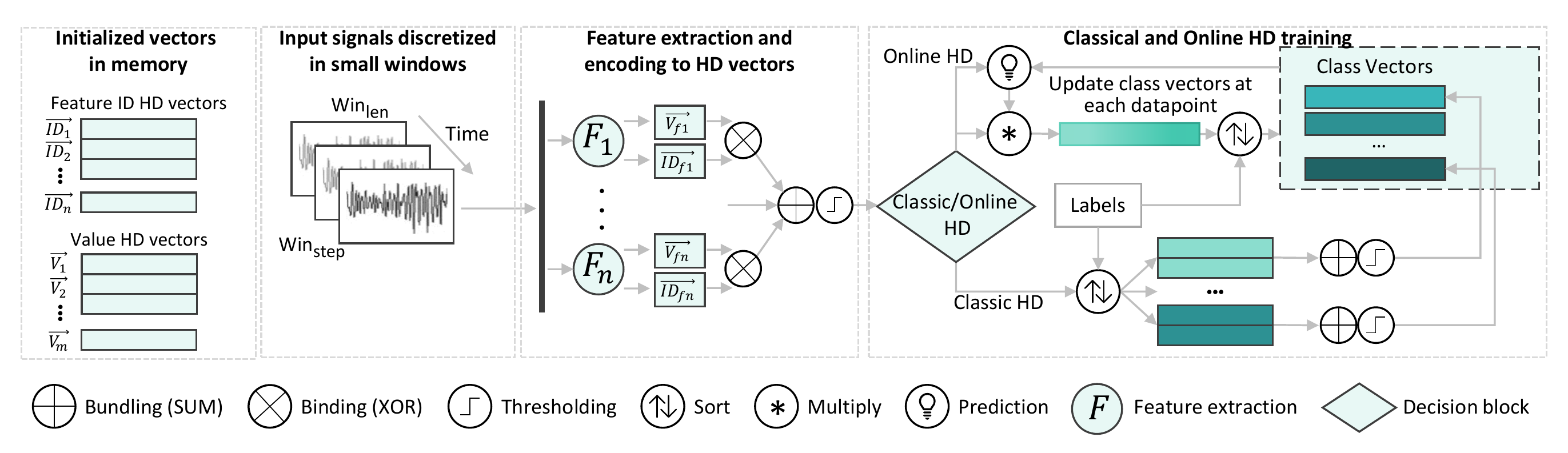}
		\caption{\small{HD workflow for training classical and online HD models. Online training differs in that the class vectors are updated after every datapoint by multiplying its similarity to the target class by the vector before accumulating it into the class.}} 
		\label{fig:hdworkflow}
	\end{figure*}

    This works contributes to the state-of-the-art in the following ways: 
    \begin{itemize}
		\item We show how HD computing models represented as hypervectors can be used to compare inter-personal seizure and non-seizure models. 
		
		\item We present several approaches to create generalized models from personalized models of individual subjects. Then, we study how many subjects are needed to achieve stable generalized models.  
  
		\item We compare the performance of epilepsy detection between personalized and generalized models, aggregated for all subjects, and also for individual subjects. Furthermore, we design hybrid models that rely both on personalized and generalized models and show that they can achieve better detection accuracy than just generalized or personalized models. 

        \item Finaly, we test knowledge transfer between models of two EEG epilepsy datasets, one big but short-term database (Repomse~\cite{rheims_hypoxemia_2019}), and one long-term but with a smaller number of subjects (CHB-MIT~\cite{shoeb_application_2009}). 
		
	\end{itemize}

	The remainder of the paper is organized as follows: Section~\ref{sec:HDcompForPersGen} motivates and details how hyperdimensional computing can be used for the creation of personalized and generalized models, how they can be compared, and how to combine them. Section~\ref{sec:Setup} describes the dataset used as well as all important methodological choices. Section~\ref{sec:Results} presents results of all analyses performed, while section~\ref{sec:Discussion} further contextualizes and comments the presented results. In the end, Section~\ref{sec:Conclusion} concludes this work.

    \section{HD computing for personalized and generalized epilepsy models} 
    \label{sec:HDcompForPersGen}
    \subsection{Basics of HD computing} 
    \label{subsec:HDbasics}

    HD computing is based on computations with long and redundant vectors (usually $\geq 10000$ dimensions), representing information in a condensed and distributed way. It was inspired by the neuroscience research hypothesis that the brain's computation is based on the high-dimensional randomized representation of data rather than scalar numerical values~\cite{kanerva_hyperdimensional_2009}.
    
      Calculating and learning with vectors (usually $\geq 10000$ dimensions) is based on two key algebraic properties: 1) any randomly chosen pair of vectors are nearly orthogonal, and 2) if we sum two or more vectors, the result will be with a high probability more similar to the added vectors than to any other randomly chosen vector~\cite{kanerva_hyperdimensional_2009}. These properties are the basis for learning and inference. 
    The most common subtype of used vectors are binary ones, where their elements can be only 0 or 1. In practice, also tertiary (-1,0,1) or integer/float vectors are sometimes used. 

    More specifically, HD learning starts by encoding data and its relations to HD vectors. As illustrated in Fig.~\ref{fig:hdworkflow} typically, baseline vectors representing different feature values and features indexes are combined during the encoding stage to get one vector. This vector represents a specific data sample, instead of a feature set, as in other ML approaches. This makes the training phase relatively simple and consists of summing (bundling) all vectors from the same class to one prototype vector representing each class. Summation of the vectors is usually done by bit-wise summation, followed by majority voting normalization. 
    Later, during inference, a vector representing the current data sample is compared with the prototype vectors of all classes, and the label of the most similar one is given as output. The similarity is measured as the distance between two vectors, which can be the Hamming distance for binary vectors or cosine (or dot) product for integer or floating-point value vectors. 

    Traditionally, HD computing classifiers have been based on single-pass learning and having a single-centroid model vector per class. Here, all data samples are equally important during learning, leading to more common patterns dominating the prototype vectors, meaning that less common patterns could be potentially under-represented and wrongly predicted even on the same training data. 
    Thus, an \textit{OnlineHD}~\cite{hernandez-cano_real-time_2021} training has been proposed, where a naive accumulation of equally important samples is replaced by using a weighting approach before adding the current vector to the prototype vectors. The weight is defined by the similarity of the current vector to the current prototype vectors: the higher the similarity, the lower the weight. This approach identifies the most dominating patterns and lowers model saturation.

    HD computing has been applied for different challenges in the domain of biomedical applications: EEG error-related potentials detection~\cite{rahimi_hyperdimensional_2020}, electromyogram (EMG), gesture recognition~\cite{rahimi_hyperdimensional_2016}, emotion recognition from GSR (galvanic-skin response), electrocardiogram (ECG) and EEG~\cite{chang_hyperdimensional_2019}, etc. 
    It has also been tested on epilepsy detection~\cite{burrello_ensemble_2021,asgarinejad_detection_2020, pale_exploration_2022}, but always with personalized models. Up to our knowledge, generalized models with HD computing have not yet been discussed.   

    \vspace{-2mm}
    \subsection{Comparing models} 
    \label{subsec:ModelsComp}
    \vspace{-2mm}
    HD computing training results in models which are in the form of long hypervectors. For example, for epilepsy, typically, we have one model (vector) for ictal (seizure, S) and one model for inter-ictal (non-seizure, NS) data. These two vectors can be easily compared using the cosine or Hamming distance, giving us the value of separability/similarity of the two classes. 
    If we train only using data from individual seizures, comparing resulting vectors can quantify the similarity between sets of seizures (intra-subject similarity). This can further help to quantify how repeatable the seizure patterns of specific patients are.
    
    Similarly, performing personalized training results in pairs of seizure (ictal) and non-seizure (interictal) model vectors for every subject. Then, an interesting next step is to study the difference between models of different subjects. This comparison leads to a characterization of the similarity of seizure patterns between different patients (ictal-ictal or S to S), the similarity of non-seizure data between patients (interictal-interictal or NS to NS), as well as inter-subject ictal-interictal (S to NS) similarity. These values can help us to understand how patients could take advantage of generalized models from other patients. Potentially, it can even help to group patients depending on the similarity between their seizure or non-seizure models. 

    \subsection{Creating generalized models} 
    \label{subsec:GenCreation}
    Focusing on generalized models, they can be created by training on the whole data containing all subjects at once, but they can also be created from already trained personalized models of individual subjects. This approach is very interesting for distributed learning where, for example, each subject's model is created on a different wearable device and without sharing sensitive clinical data. Generalized models can be combined on a central server where all individual patient models are sent. It is very important to highlight that these personalized models are privacy-preserving, as it is impossible to decode back individual data after it was encoded and accumulated to hyperdimensional models. 
    
    We test and compare different approaches to build general models from personalized ones: 1)~\textit{Avrg}, 2)~\textit{WSub} and 3)~\textit{WAdd\&Sub}.  The simplest \textit{Avrg} approach is based on the average of all personalized model vectors (and normalize to have binary model vectors again), as defined in equation~\ref{eq_Avrg}. $VecP$ stands for personalized vector of individual subject (weather it is seizure or non-seizure model vector), and $NumS$ stands for number of subjects that are available in a dataset. 
    Another approach to creating generalized models is similar to \textit{OnlineHD} training, where before adding a new subject, its model vectors are compared to the current generalized models and added proportionally to the novelty they would bring. More specifically, the personal vector of the correct class is added, but the opposite class is also subtracted and multiplied with the weight. This is defined in equation~\ref{eq_Wsub}. We call this approach \textit{Wsub} as only the opposite class is multiplied with the weight. The third tested approach is called \textit{Wadd\&sub} as defined by equation~\ref{eq_Waddsub}. Here the correct class is also multiplied with the corresponding weight before being added to the generalized model vector. In the end, we also test if there is any benefit of performing weighted adding in an iterative manner. In all cases, the weight for adding the correct class is defined as  
    \begin{equation}\label{eq_wcorr}  w_{corr} =\alpha ( 1 - hammDist_{corr})  \end{equation}
    so that the more similar (higher hamming distance) personalized and generalized vectors are, the smaller is the weight with which it is multiplied, as it doesn't bring a lot of new information. 
    On the other hand, weight for the opposite/wrong class is defined as
    \begin{equation}\label{eq_wwrong} w_{wrong} =\alpha  * hammDist_{wrong} \end{equation}
    so that if the opposite class personal vector is very similar to the current generalized vector (hamming distance is high), weight is also high, reducing that vector significantly. 
    In both equations for weight, there are additional factors $\alpha_{corr}$ and $\alpha_{wrong}$ that define the importance of adding the correct class or subtracting the wrong class. In this case, we use value 1 for both of them, but they could be further optimized in the future, depending on the specific use case.  

    \begingroup
    \footnotesize
    \begin{equation}\label{eq_Avrg}
    VecG = \frac{\sum_{p}^{NS} {VecP_{corr}  }  } {NS}  
    \end{equation}

    \begin{equation}\label{eq_Wsub}
    VecG = \frac {  \sum_{p}^{NS} {VecP_{corr} -  w_{wrong} VecP_{wrong}} }   {\sum_{p}^{NS} {1 -  w_{wrong} }   }  
    \end{equation}

    \begin{equation}\label{eq_Waddsub}
    VecG = \frac{\sum_{p}^{NS} {w_{corr} VecP_{corr} -  w_{wrong} VecP_{wrong}}}  {\sum_{p}^{NS} {w_{corr}  -  w_{wrong} }   }  
    \end{equation}
    \endgroup


    
    Further, we can also track the evolution of generalized models as more subjects are added. For example, tracking average similarity with personalized models as we add more personalized models enables us to determine how many subjects are needed to reach stable generalized models that do not change significantly when adding more subjects. 

    \subsection{Hybrid models} 
    \label{subsec:HybridModels}
    Using generalized models would be most convenient from a practical perspective. Models could be trained once and then used on all new subjects. This is particularly interesting for wearable outpatient applications. But as epilepsy exhibits highly personalized patterns, generalized models usually do not perform well enough. Also, real-life recordings are highly unbalanced, which poses another challenge for good performance, as shown in~\cite{pale_multi-centroid_2022}. 

    Having both personalized and generalized models, we can compare the detection performance on individual subjects. We can further divide patients into groups for which personalized perform better, or for which generalized models are better. 
    Here we investigate the hybrid approach of using generalized models by default and replacing them with personalized models if the generalized performs worse than a certain threshold. We study the percentage of subjects that would need to use personalized models depending on the chosen performance threshold of generalized and how the overall performance over the whole population changes when including more subjects using generalized model. 
    
    Finally, HDC models allow us to have even more flexibility, for example, by combining both personalized and generalized models at the same time, rather than choosing only a personalized or a generalized model. More specifically, since seizure and non-seizure models are just simple hypervectors, we test the performance when only the seizure model is kept personalized, while the non-seizure model vector is kept generalized (\textit{NSgen-Spers}), or the other way around (\textit{NSpers-Sgen}). 
    The \textit{NSgen-Spers} approach is interesting as it reuses non-seizure models from other subjects where only seizure data from individual subjects are needed to create their personal seizure models (\textit{NSgen-Spers}), potentially saving a lot of time for training personalized models. 
    On the other hand, the \textit{NSpers-Sgen} model is very interesting since we only require non-seizure data of each subject to have a personalized model, which is much easier to gather than seizure data. This would mean that the new subject would only need to record some amount of non-seizure data rather than recording for days or weeks to gather sufficient amount of seizure data. In this work, we compare the performance of both of these approaches with personalized and generalized model performances.

    \begin{figure}
		\centering
		\includegraphics[width=\linewidth]{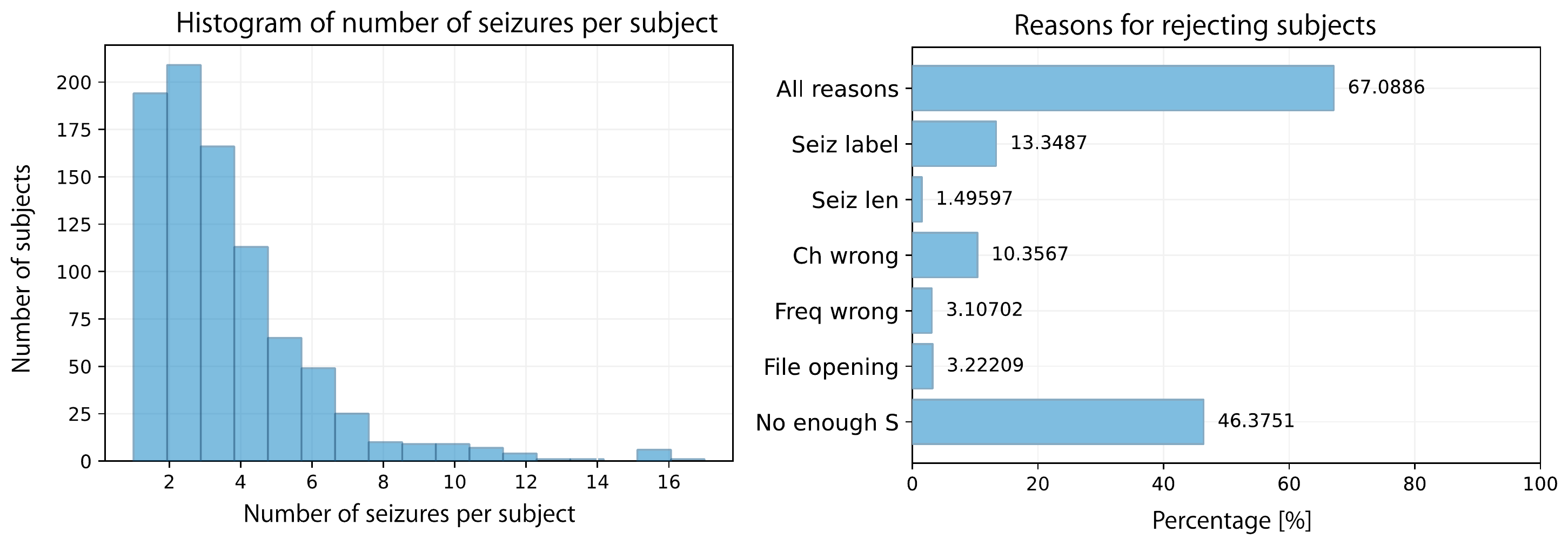}
        \vspace{-4mm}
		\caption{\small{ Statistics of the Repomse dataset: number of seizures per patient in original database, and also reasons for rejecting specific subjects. In the end we keep 286 subjects which satisfy our criteria. } } 
		\label{fig:RepomseStats}
	\end{figure}

    \subsection{Models transfer between databases} 
    \label{subsec:Transdatabases}
    In the literature, knowledge transfer between models of different datasets is rarely tested. In the recent paper~\cite{amirshahi_m2d2_2022}, deep learning approach called M2D2 ('Maximum-Mean-Discrepancy-Decoder') was used for automatic temporal localization and labeling of seizures in long EEG recordings of the Epilepsiae~\cite{klatt_epilepsiae_2012} and CHB-MIT databases. Their problem formulation is slightly different than ours, detecting windows of a certain size with a high probability that they contain seizures, intending to simplify and speed up the labeling process of a new database. But overall, this is one of the few works that test models trained on a different dataset.
    
    For hyperdimensional computing, based on our knowledge, there is no work that performed generalized training so far, or that further combined models from several datasets. Here, HD computing enables us not only to use generalized models from different datasets but also to create hybrid models where only one part of the model is from a different dataset. For example, we will test when only the seizure (or non-seizure) model is from a different dataset. This enables us to better understand the knowledge transfer between models of two datasets, in both directions.

    \section{Experimental setup} 
    \label{sec:Setup}

    \subsection{Dataset}
     The main analyses were performed using the Repomse database~\cite{rheims_hypoxemia_2019} curated by the Lausanne University Hospital. Originally, the database contained EEG recordings of 869 patients with an average of 3.3 seizures per patient. A more detailed distribution of the number of seizures per subject is shown in Fig.~\ref{fig:RepomseStats}. We selected only patients with at least three seizures (to be able to create personalized models), with identical 18 channels (i.e., FP1-F7, F7-T7, T7-P7, P7-O1, FP1-F3, F3-C3, C3-P3, P3-O1, FP2-F4, F4-C4, C4-P4, P4-O2, FP2-F8, F8-T8, T8-P8, P8-O2, FZ-CZ, CZ-PZ) and with a sampling frequency of at least 256Hz. If the sampling frequency was higher, it was downsampled to 256Hz. Fig.~\ref{fig:RepomseStats} also shows the reasons for rejecting specific subjects and the percentage of subjects affected. Finally, we kept 286 subjects, which is still a much larger number than we could have get from any other publicly available dataset, and then it was ideal for studying generalized models. Original recordings contain a single seizure with a maximal length of 1 min,  with usually 3 minutes of interictal recording before the seizure onset. Similarly, as in~\cite{gabeff_interpreting_2021} we exclude 1 min of pre-ictal, and utilize 1 min of inter-ictal data. In this way, every file contains a balanced amount of seizure and non-seizure data. 
     
    Finally, to test knowledge and models transfer between databases, we also used the public CHB-MIT database~\cite{shoeb_application_2009}. It is a long-term scalp EEG database for epilepsy detection containing 980 hours of data recorded at 256Hz. It consists of 183 seizures from 24 pediatric subjects. On average, it has 7.6 $\pm$ 5.8 seizures per subject, and between 23 and 26 channels, of which the same 18 channels as in Repomse database are kept. 
    From the raw database, similarly as in~\cite{pale_exploration_2022} we prepared a dataset that contains ten times more non-seizure data than seizure data to be closer to a more real-life data balance. The main reason to avoid a balanced scenario, which is common in many works in the literature, is that it can lead to a highly overestimated performance, not achievable during the continuous monitoring with a wearable device~\cite{pale_multi-centroid_2022}.
    Non-seizure segments were chosen randomly from available non-seizure data, but excluding data 1 min before and 15 min after a seizure, and arranged around seizures. Thus data structure was similar as in Repomse, with one-seizure-per-file but with a bigger amount (10x more) of non-seizure data.

    \subsection{Features used}
    We use the mean amplitude and both relative and absolute power spectral density in the five common brain wave frequency bands: delta: [0.5-4] Hz, theta: [4-8] Hz, alpha: [8-12] Hz, beta: [12-30] Hz, gamma: [30-45] Hz, and low-frequency components: [0-0.5] Hz and [0.1-0.5] Hz. We also included the line length feature ~\cite{esteller_line_2001} and the 'approximate-zero-crossing' features (AZC), as proposed in~\cite{zanetti_approximate_2022}. The AZC features are based on simplifying the EEG signals by applying a polygonal approximation to mimic how our brain selects prominent patterns among noisy data. Then, a simple zero-crossing count is used as an estimation of the dominating frequency of the signal. We extracted six features by using six different thresholds for polygonal approximation. 
    Thus, in total, we extract 25 features from data segmented into 4-second windows with a 0.5-second step.  
    Before extracting the AZC features, the data is filtered with a 4th-order, zero-phase Butterworth bandpass filter between [1, 20] Hz.

    \begin{figure}
		\centering
		\includegraphics[width=\linewidth]{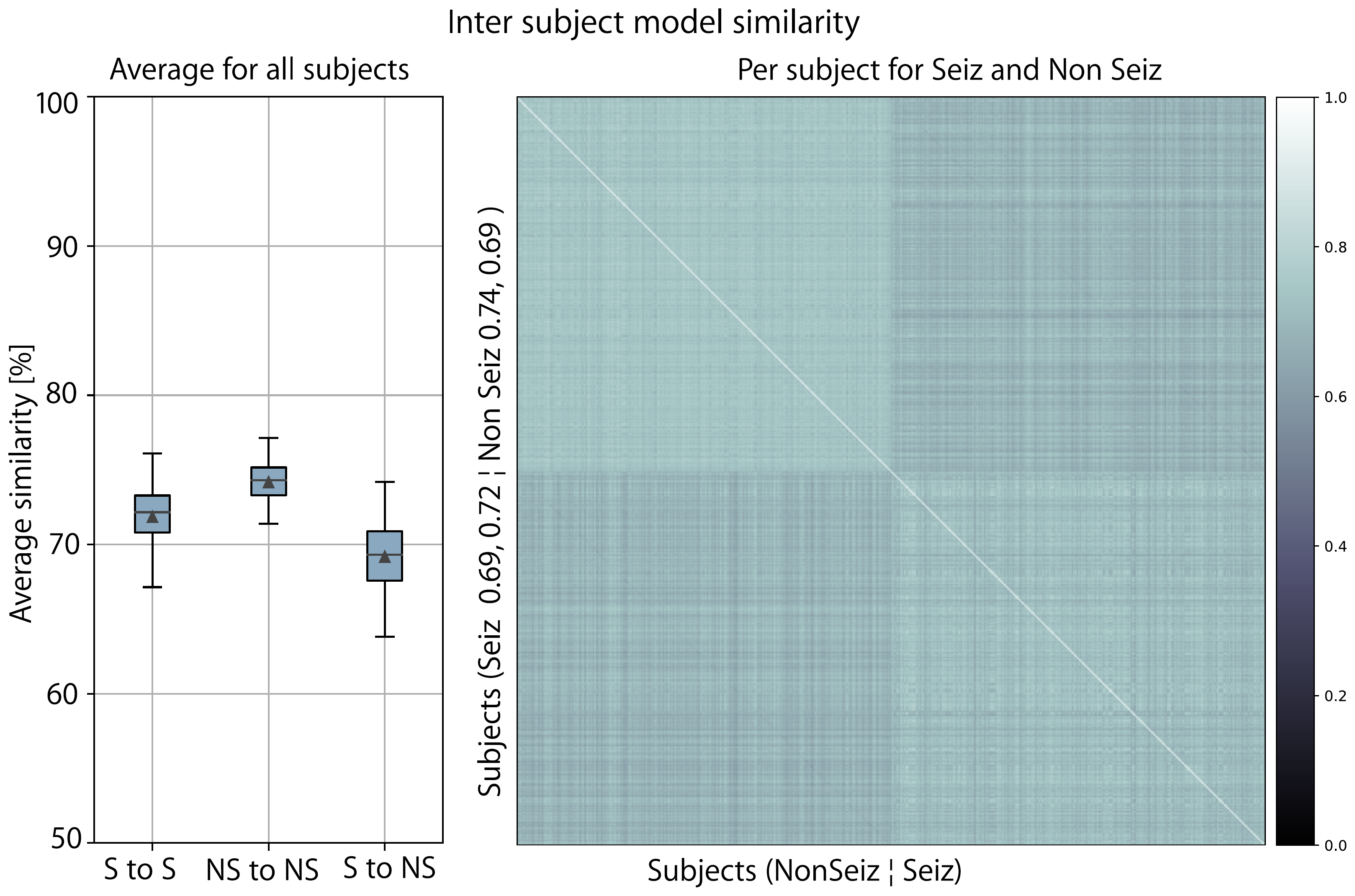}
        \vspace{-4mm}
		\caption{\small{ Inter-subject similarity between model vectors of individual subjects (both between ictal (Seiz, S) or inter-ictal (NonSeiz, NS) models). } } 
		\label{fig:InterIntraSimilarity}
	\end{figure}

    \subsection{Training and evaluation}
    For the HD computing workflow, we can choose between the standard learning or \textit{OnlineHD} approach. Since in~\cite{pale_exploration_2022} authors compared standard and \textit{OnlineHD} for the use case of epileptic seizure detection and showed better performance of an \textit{OnlineHD}, we use it in this work as well.
    Due to the subject-specific nature of epileptic seizures, and one-seizure-per-file division, training and evaluation of personalized models were done using leave-one-seizure-out cross-validation. In the end, all test predictions are appended into the original time order, and performance is calculated. This is a training method that is more appropriate for long-term datasets, without excluding any data and thus leading to more representative performance for real-life applications~\cite{simon_hdtorch_2022}.
    For generalized models, on the other hand, and for both databases, the leave-one-subject-out cross-validation approach was used. To report the overall performance, we calculate the average over all subjects, both for personalized and generalized approaches. 
   
    The performance of the classifier is evaluated with respect to episode detection and duration-based detection, measuring sensitivity, predictivity, and F1 score. The performance at the episode level groups the signal into blocks of seizure and non-seizure, as detailed in~\cite{pale_importance_2023}, while the performance at the duration level considers the correct prediction of each sample. 
    In epilepsy detection, raw label predictions often lead to unrealistic seizure dynamics (e.g., seizures lasting only a few seconds or seizures that are only a few seconds apart). Thus, label post-processing is an integral part of the pipeline. 
    We use Bayesian post-processing that calculates cumulative probability of classes and thresholds them as described in~\cite{gabeff_interpreting_2021}. Threshold value of 1.5 and window length of 5s were used.

    \section{Results} 
    \label{sec:Results}
    \vspace{-2mm}
    First, we show how HD computing can compare seizure and non-seizure models of individual subjects. Then we show results related to the creation of generalized models, followed by a comparison of the performance of personalized and generalized models. Finally, we show the possibilities of hybrid models and models trained on different datasets. 

    \begin{figure}
		\centering
		\includegraphics[width=\linewidth]{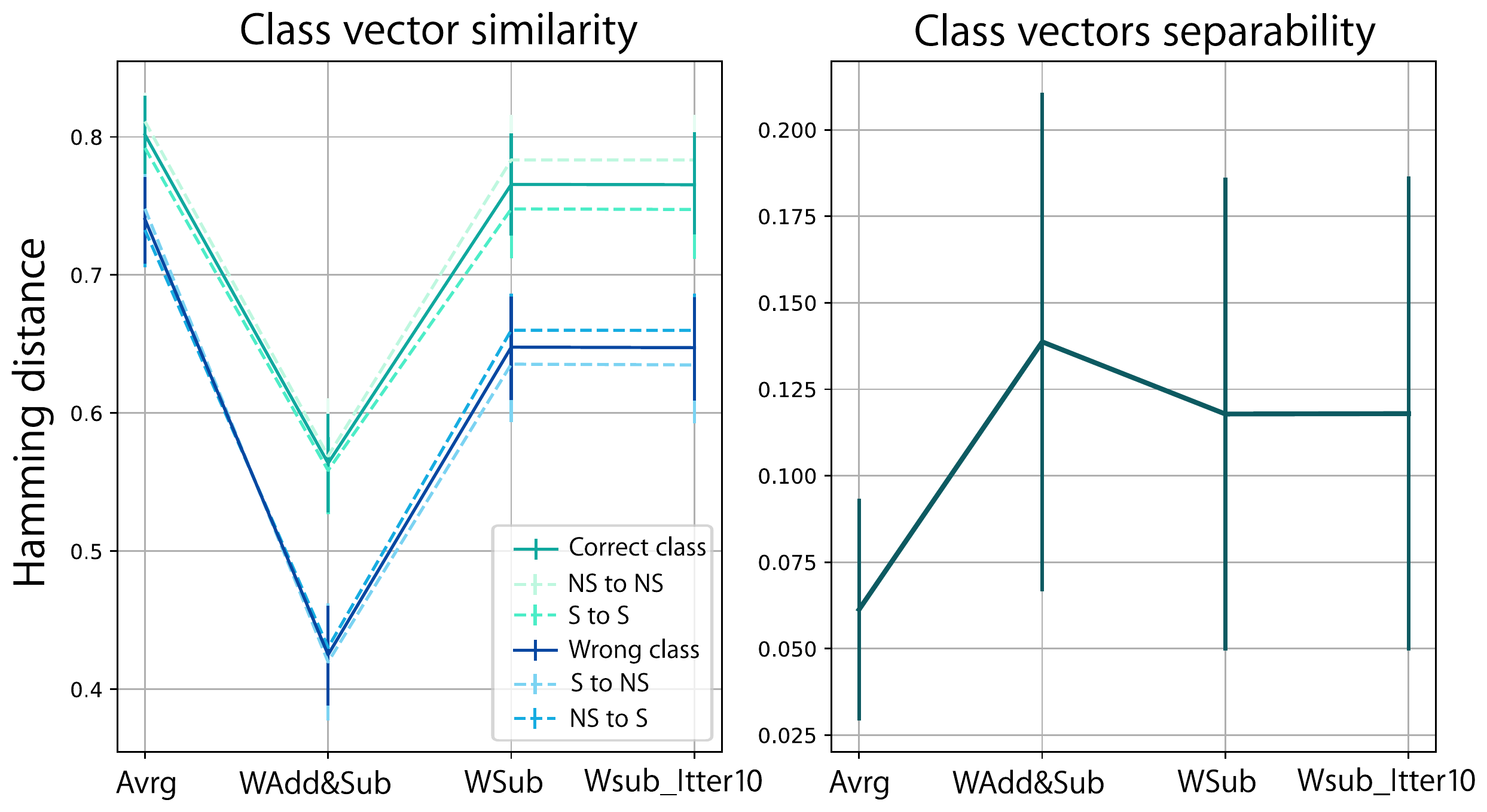}
        \vspace{-4mm}
		\caption{\small{ Comparing different approaches to creating generalized models from personalized. Similarity between seizure and non-seizure model vectors as well as overall class separability is measured. } } 
		\label{fig:GeneralizedCreationMethod}
	\end{figure}

    \subsection{Inter-subject similarity} 
    \label{subsec:ModelSimilarityRes}
    \vspace{-2mm}
    Fig.~\ref{fig:InterIntraSimilarity} shows a comparison between individual seizure and non-seizure models of 286 subjects. Non-seizure to non-seizure (NS to NS) models are the most similar, followed by seizure to seizure (S to S). Finally, seizure to non-seizure (S to NS) models are the least similar, which is expected. There is significant difference between \textit{S to S} and \textit{NS to NS}, and also there is a significant difference between \textit{S to NS} similarities and \textit{S to S} as well as \textit{NS to NS}. Wilcoxon paired tests were performed to compare distributions, with all $p-$values being lower than $10^{-15}$. 
    A similar comparison can be made within the data of one single subject (\textit{intra-subject}), comparing models of individual seizures. 
    
    \subsection{Creating generalized models} 
    \label{subsec:GenCreationRes}
    \vspace{-2mm}
    Fig.~\ref{fig:GeneralizedCreationMethod} shows the quality of generalized models created with four different approaches. We measure the similarity between the generalized and all individual personalized models, showing the mean and standard deviation. The similarity is measured for all options: generalized seizure to personalized seizure models (\textit{S to S}), non-seizure to non-seizure (\textit{NS to NS}), which represent similarities between correct classes, and similarly \textit{NS to S} and \textit{S to NS}, representing similarities between opposite classes. From similarities between correct and opposite classes, we further calculate class vector separability as a measure of how separable the models are.
    
    When comparing the four approaches, the simplest averaging (\textit{Avrg}) of individual subject models results in the highest similarity between the correct classes but also a high similarity with the opposite class, leading to the lowest class separability. When we add weighted subtraction of opposite class vectors in the \textit{Wsub} approach, both similarities drop, but opposite classes become less similar, leading to better separability. Finally, if adding the correct class individual vectors is also multiplied with weight (proportional to the novelty of individual subjects vector) accompanied with weighted subtraction of opposite class (\textit{WAdd\&Sub}), similarities drop significantly but lead to the highest classes separability. 
    Since the order in which subjects were added could have influenced the similarity of generalized models with personalized ones that were added at the beginning of the process, we also tested iterative creation, where we passed through all subjects several times. As shown in Fig.~\ref{fig:GeneralizedCreationMethod} this didn't change any of the measured values.
    
    \begin{figure}
		\centering
		\includegraphics[width=\linewidth]{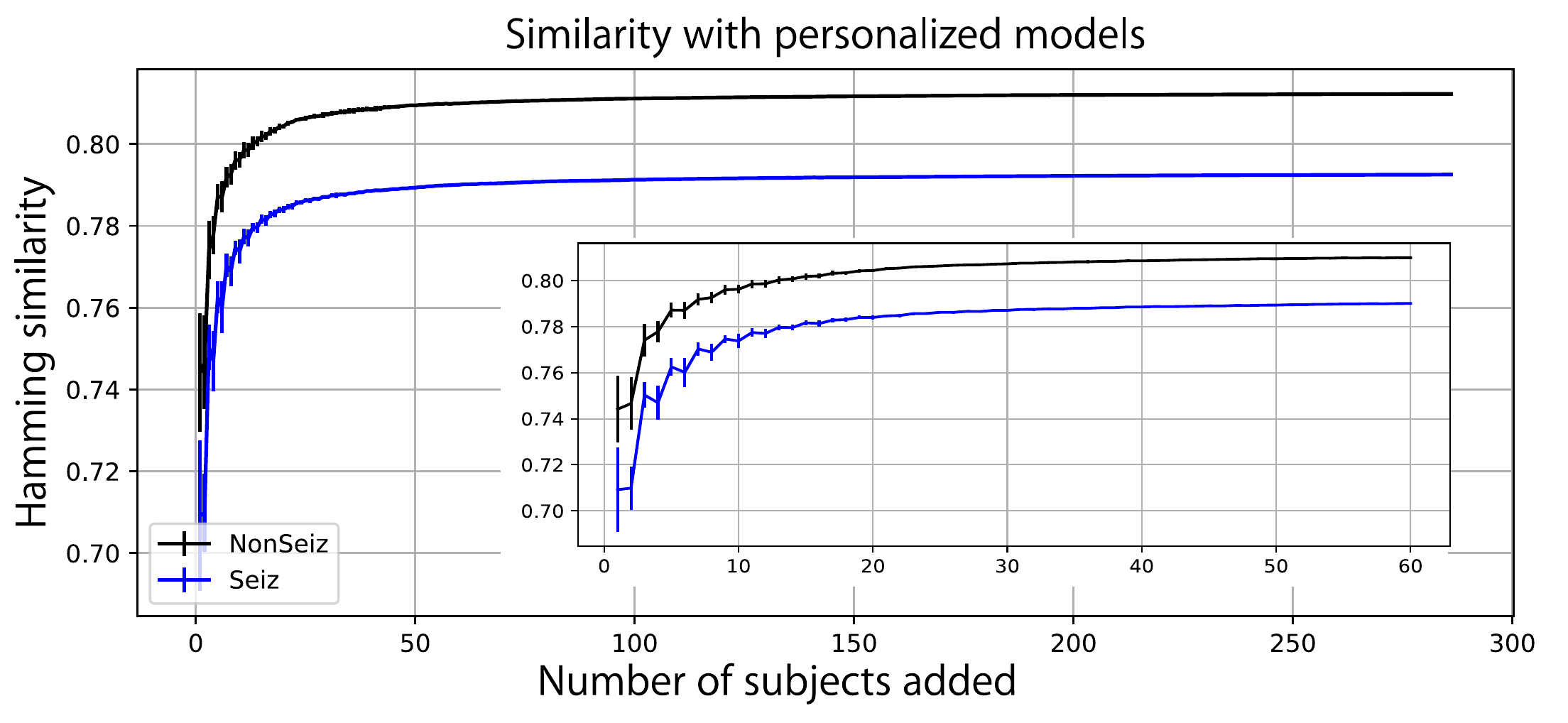}
        \vspace{-4mm}
		\caption{\small{ Evolution of generalized vectors as adding one by one individual subject. Average similarity with all personalized subjects is measured to characterize how stable are generalized vectors. } } 
		\label{fig:GenGeneration}
    \vspace{2mm}
	\end{figure}

    Next, we tried to assess how many individual subjects are needed to get stable generalized models. Thus, we measure how similar the generalized model is on average to all individual personalized ones (no matter if they were already added to the generalized or not). We repeated this 10 times, adding subjects in a different order, and we show the average similarity in Fig.~\ref{fig:GenGeneration}. A clear trend of similarity plateauing after adding approximately 40 patients (both for seizure and non-seizure vectors) quantifies the minimum number of patients needed to create stable generalized models.

    \subsection{Personalized vs. Generalized models} 
    \label{subsecPersVsGen}
\begin{figure}
		\centering
		\includegraphics[width=\linewidth]{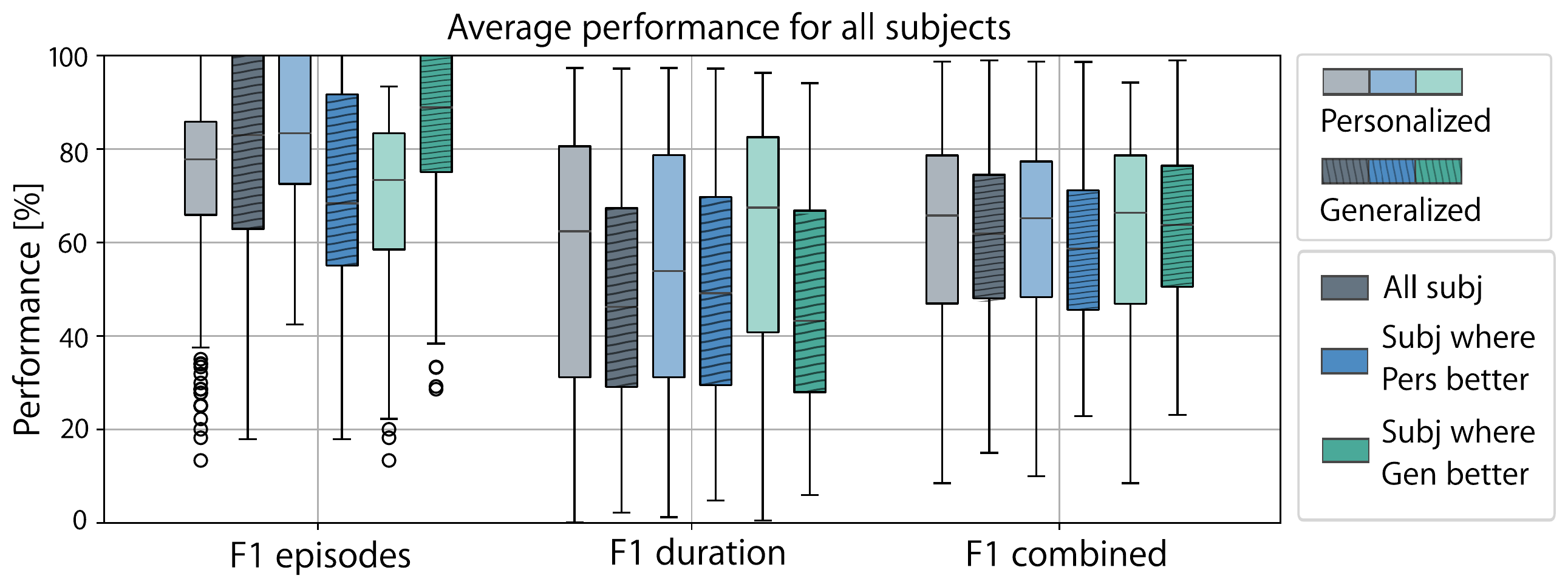}
        \vspace{-4mm}
		\caption{\small{ Comparing performance of personalized and generalized models: 1) for all subjects (in gray), 2) on subjects where personalized models perform better (in blue) and 3) on subjects where generalized perform better (in green). Performance without any postprocessing is shown, for F1 score on the level of episodes, duration and gmean of both (F1 combined).  } } 
		\label{fig:PersGenStratification}
    \vspace{2mm}
	\end{figure}

    First, we were interested in comparing performance between personalized and generalized models. If we look at the F1 score for episodes, personalized models were better for 95 out of 286 subjects, generalized were better for 152 subjects, and for 39 subjects, both models had the same result. In Fig.~\ref{fig:PersGenStratification}, we compared the performance of generalized and personalized models in three cases: the average of all subjects, the average of subjects where generalized models performed better (F1 score for episodes), and the average of subjects where personalized models performed better. 
    The figure shows that for the whole group population, generalized performed, on average, slightly better than personalized models in terms of F1 score for episodes (F1E), but when looking at the F1 score for the duration (F1D), where we care about the classification of each sample, personalized models performed much better. When observing only subjects where personalized performed better, it can be seen that, indeed, F1 score for episodes (F1E) dropped significantly for generalized performance. Similarly, for subjects where generalized performed better, F1E was much higher for generalized than personalized models. This gap between performance of personalized models and generalized models for individual subjects opens the possibility to stratify patients into two groups. 
    
\begin{figure}
		\centering
		\includegraphics[width=\linewidth]{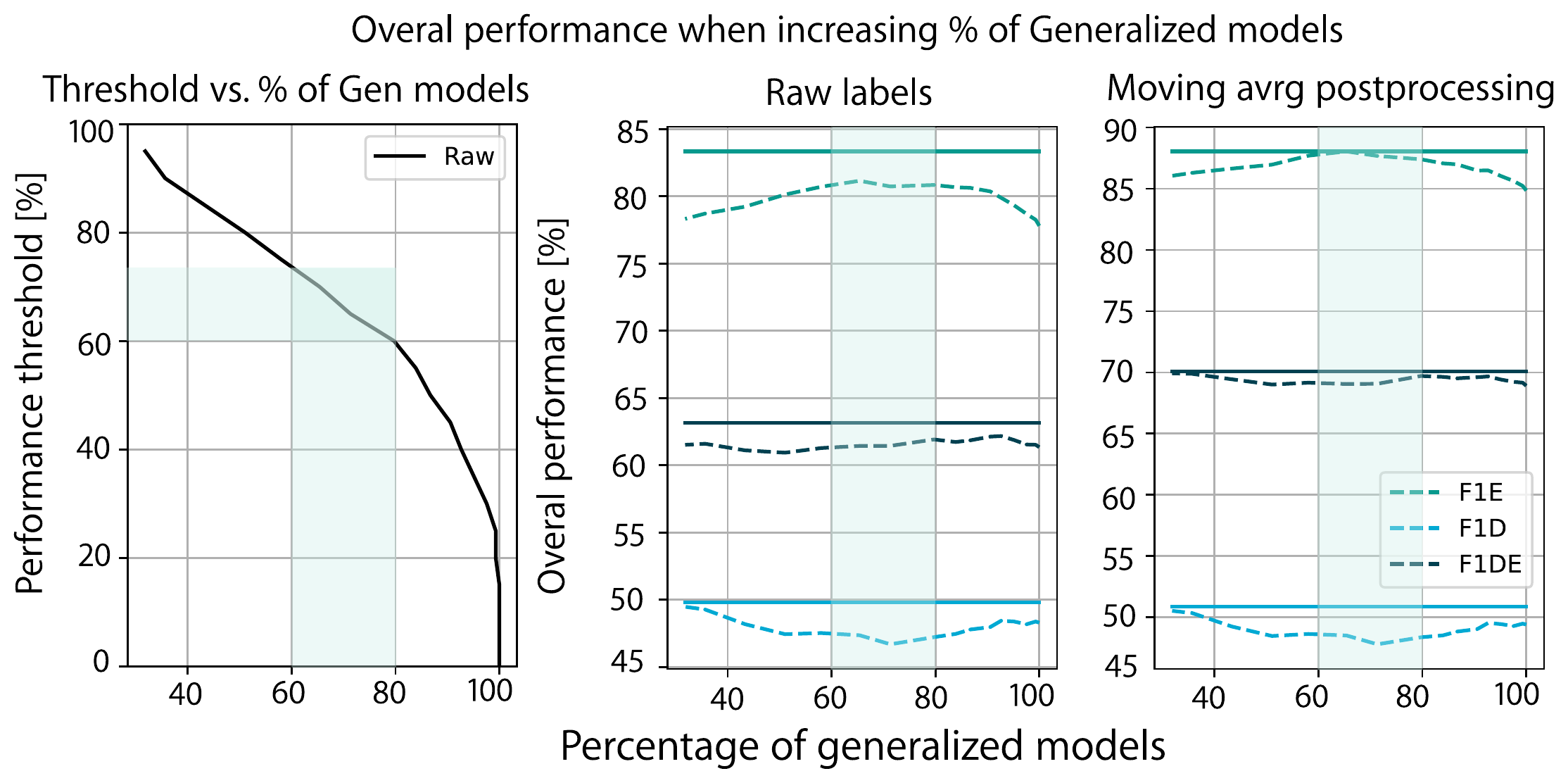}
        \vspace{-4mm}
		\caption{\small{ We investigate combinations of generalized and personalized models. Namely, by default for all subjects generalized models is used, but if it performs worse then certain performance threshold, personalized models is used. On the left graph percentage of subjects that would use generalized model is shown as threshold is decreased. On next two graphs, overall performance when changing percentage of generalized models is shown.  } } 
		\label{fig:PersGenAddingMoreGen}
	\end{figure}

    As explained before, for real-life applications, ideally one generalized model that would perform well for all subjects would be the best solution. But as shown, generalized models do not always perform the best. Thus, we tested an approach in which we used personalized models in case the generalized model performed under a chosen threshold. We measured how overall performance for all subjects depends on the chosen threshold and the final percentage of generalized models used. In Fig.~\ref{fig:PersGenAddingMoreGen}, results show that the highest overall performance (average of all subjects in the Repomse database, if looking at F1E) is achieved when 60 to 80 \% of subjects used generalized models. With full horizontal lines, the hypothetical performance with an optimal selection of personalized or generalized models among subjects. This is of course not achievable in practice, because we cannot know which model performs better if both models are not built and tested. Thus here we use approach where by default generalized model is used, unless it performs worse then chosen threshold. More precisely, if the minimal satisfactory performance threshold of generalized models is set to 60\% this would result in 80\% of patients having generalized models. If we want to be more strict and choose a performance threshold of 75\% we will use 60\% of generalized models. 
    On the right graph in Fig.~\ref{fig:PersGenAddingMoreGen}, we show the performance after moving average post-processing, and we can see that even if for raw predictions the ideal performance (full horizontal lines) were not reached, with post-processing it is possible to reach it. 

    \subsection{Hybrid models} 
    \label{subsec:HybridModelsRes}
\begin{figure}
		\centering
		\includegraphics[width=\linewidth]{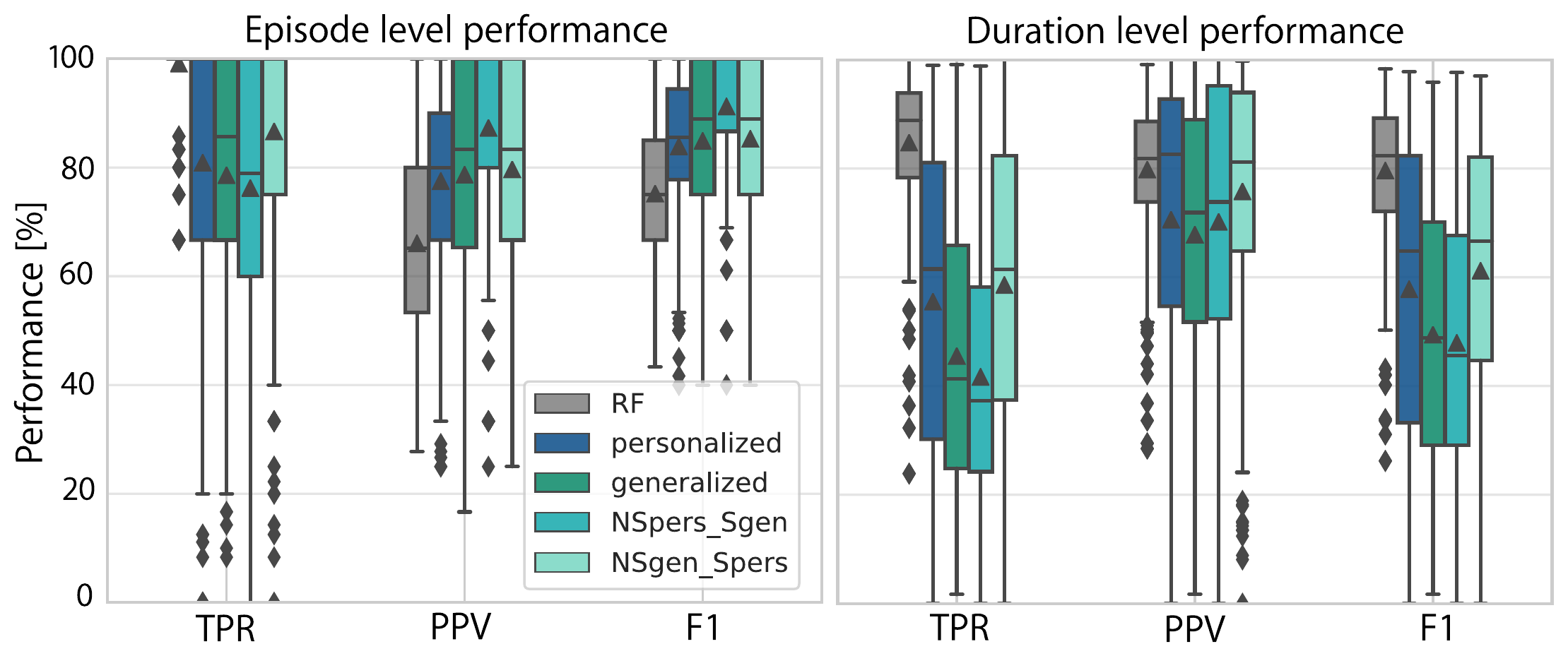}
        \vspace{-4mm}
		\caption{  Performance of 4 types of models: personalized, generalized and hybrid  (\textit{NSgen-Spers} and \textit{NSpers-Sgen}). Boxplots represent performance over all subjects. Sensitivity (TPR), precision(PPV) and F1 score both for performance on the level of episodes and on the whole duration level are shown. } 
		\label{fig:All4modelsPerformance}
	\end{figure}

    Finally, a unique possibility of HD computing models is to actually not only choose a personalized or generalized model per subject, but the ability to combine and create hybrid models. Fig.~\ref{fig:All4modelsPerformance} shows the performance of random fores, personalized, generalized and two hybrid models (\textit{NSgen-Spers} and \textit{NSpers-Sgen}). Sensitivity (TPR), precision (PPV) and F1 score for episode and duration based performance are shown. 
    
    When looking at episode performance, HD computing models outperformed random forest with the same set of features. Further, generalized and personalized models perform similarly, and hybrid \textit{NSpers-Sgen} models achieves slightly higher F1 score for episodes. This is due to higher precision (PPV), or less false positives. If we look at performance on the level of duration, it can be noticed that generalized and \textit{NSpers-Sgen}  models perform significantly worse in terms of sensitivity (TPR), leading also to a lower final F1 score. The \textit{NSgen-Spers}  approach, on the other hand, performs as well as personalized models in terms of sensitivity, having also a slightly higher precision and finally a slightly higher final F1 score (F1D) than personalized models. This means that non-seizure data can be used from other subjects but that for good sample-by-sample classification it is important to keep seizure models personalized.

    \vspace{-2mm}
    \subsection{Models transfer between databases} 
    \label{subsec:TransdatabasesRes}
\begin{figure}
		\centering
		\includegraphics[width=\linewidth]{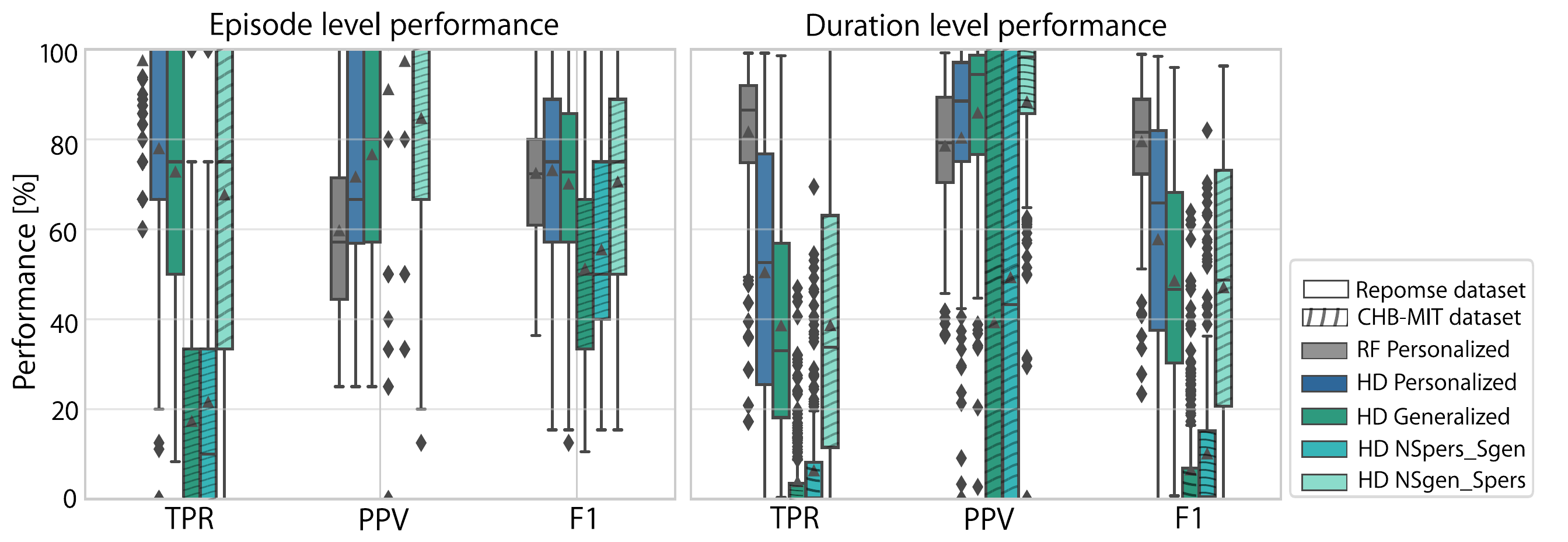}
        \vspace{-4mm}
		\caption{ Knowledge transfer between two databases is tested on the Repomse dataset. Performance of 5 types of models is shown: personalized, generalized Repomse (models from Repomse data), generalized CHB-MIT, and hybrid from CHB-MIT  (\textit{NSgen-Spers} and \textit{NSpers-Sgen}).
      } 
		\label{fig:TransDataset_Repomse}
	\end{figure}

    \vspace{-2mm}
   Fig.~\ref{fig:TransDataset_Repomse} shows the result of testing models trained on one database (CHB-MIT) and utilizing (testing) them on a different database (Repomse). More specifically, the performance of five different models is shown: personalized models and generalized models trained on Repomse, followed by three models trained on CHB-MIT database: generalized and hybrid (\textit{NSgen-Spers} and \textit{NSpers-Sgen}). Boxplots represent the performance distribution over all subjects. First, a clear drop in F1 score (for episode and duration) when using a generalized model from the CHB-MIT dataset is visible. This is due to much worse seizure detection as measured by significantly lower sensitivity (TPR) both on episode and duration levels. 
    The \textit{NSpers-Sgen} approach doesn't help to improve performance, but the \textit{NSgen-Spers} does help. Indeed, personalizing seizure models improves sensitivity and also the final F1 score (both for episodes and duration), reaching and even exceeding the levels of the generalized model from the same Repomse database.  

\begin{figure}
		\centering
		\includegraphics[width=\linewidth]{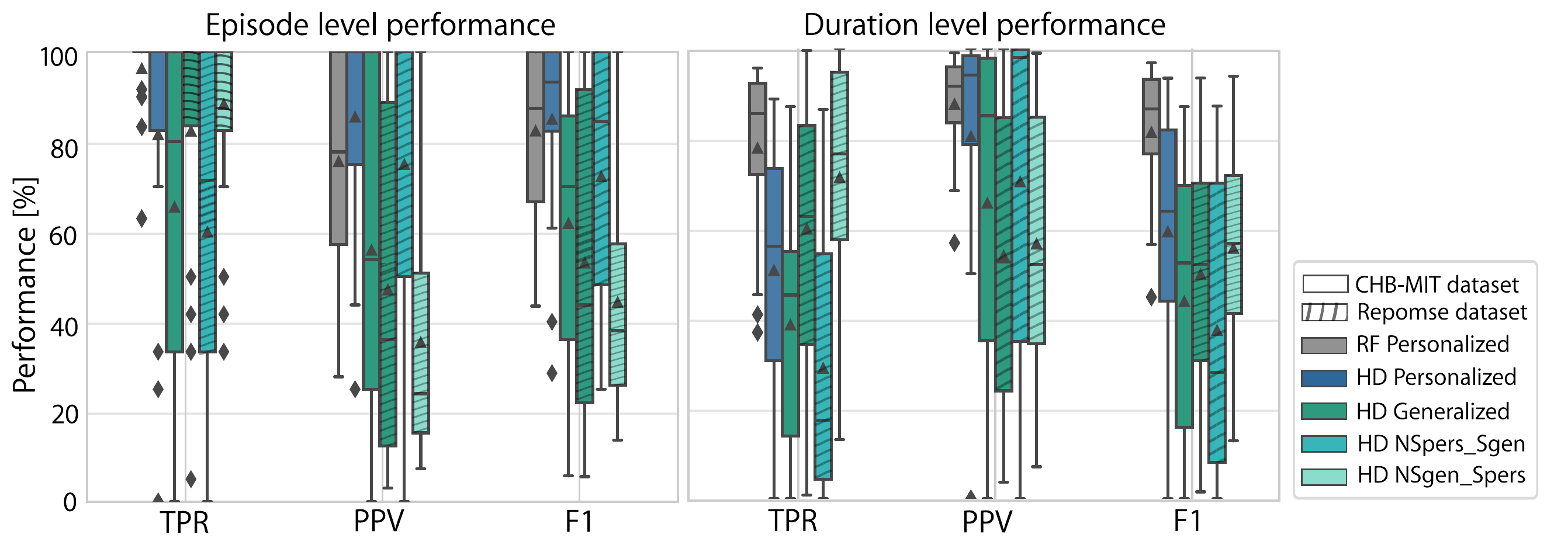}
        \vspace{-4mm}
		\caption{ Knowledge transfer between two databases is tested on the CHB-MIT  dataset. Performance of 5 types of models is shown: personalized, generalized CHB-MIT (models from CHB-MIT data), generalized Repomse,  and hybrid from Repomse  (\textit{NSgen-Spers} and \textit{NSpers-Sgen}). } 
		\label{fig:TransDataset_CHBMIT}
	\end{figure}

    Fig.~\ref{fig:TransDataset_CHBMIT} shows the opposite transfer of models. Models trained on Repomse dataset with many subjects are tested on the CHB-MIT dataset. Here \textit{NSpers-Sgen} models performed better, as in Repomse there is a very small amount of non-seizure data, and thus it doesn't generalize well for a dataset with much more non-seizure data. On the other side, Repomse contained more subjects and thus seizure data seems to be general enough to use generalized models.

    \vspace{-2mm}
    \section{Discussion} 
    \label{sec:Discussion}
    \vspace{-2mm}
    
    By analyzing inter-patient similarities, we show that non-seizure models are more similar than seizure models. This makes sense, as seizures usually exhibit specific individual patterns. Still, seizures to non-seizures are the least similar, enabling classification. For long-term, unbalanced epilepsy recordings, these trends would probably be even stronger, especially regarding non-seizure models. 
    Further, this opens the possibility of checking similarities between individual patients and potentially grouping them into groups of similar seizure (or non-seizure) patterns. 
    
    Furthermore, we studied the creation of generalized models by adding personalized models one-by-one in three different ways. It is shown that simple averaging of personalized models leads to the most similar vectors to the summed personalized ones, but also leads to low class separability. Approaches where vectors are weighted before being added,  and also the idea of subtracting opposite class, leads to more separable generalized vectors. Thus, we recommend using one of those approaches for creating generalized model vectors. Iterative passing through subjects doesn't lead to better-generalized models. This is most likely due to the fact that we used 286 subjects, and generalized models were very stable, so passing again through individual models didn't change generalized models anymore. 
    Assessment of the evolution of generalized models as, one-by-one, more individual subject models were added, lead to the conclusion that at least 40 individual subjects are needed to get stable generalized model vectors. We have to consider that different databases with unbalanced datasets result in different minimal number of subjects. For example, if data is long-term (such as, for example, in CHB-MIT epilepsy database), maybe non-seizure vectors will stabilize much sooner, whereas, for stable seizure models, we will need many more subjects. 

    Assessment of the evolution of generalized models as, one-by-one, more individual subject models were added, lead to the conclusion that at least 40 individual subjects are needed to get stable generalized model vectors. We have to consider that different databases with unbalanced datasets result in different minimal number of subjects. For example, if data is long-term (such as, for example, in CHB-MIT epilepsy database), maybe non-seizure vectors will stabilize much sooner, whereas, for stable seizure models, we will need much more subjects. 
    
    Observing the performance of personalized and generalized models led to the conclusion that there is an obvious stratification between subjects. This could be due to not enough seizures or very different seizures in some patients, which makes personalized models not precise enough. It can also be that some subjects have very specific seizures for which generalized seizure models don't perform well.      
   
    Even if utilizing generalized models for all subjects would be the most practical, generalized models do not perform well for all subjects. In Fig.~\ref{fig:PersGenAddingMoreGen}, we can see that when caring about sample-by-sample prediction, personalized models always perform better. With a simple system where personalized models are built only for subjects where generalized models performed worse than some preset threshold, it is possible to achieve even better performance than with only generalized (or even only personalized models). Specifically, for the Repomse dataset, it is possible to maintain generalized models for up to 80\% of patients without loss of performance. This is probably dataset-specific, and numbers might be significantly different if long-term datasets are used. 

    Finally, we tested the detection performance of hybrid models, containing one generalized and one personalized model vector (\textit{NSgen-Spers} and \textit{NSpers-Sgen}), and show that, indeed, they can improve overall seizure detection even more. For example, if we require high episode performance, the \textit{NSpers-Sgen} model gives the best overall performance by decreasing the number of false positive seizure predictions. In contrast, if we care that the entire duration of the seizure is correctly classified, the (\textit{NSgen-Spers} model performs better by keeping high sensitivity. 
    We tested the possibility of keeping both personalized and generalized models for each subject and giving a prediction of a more certain model, but this led to giving a prediction of the personalized model in almost all the cases, thus not being particularly interesting.

    When comparing the possibility of transferring knowledge between databases, interesting results were achieved. Transferring knowledge from the long-term CHB-MIT database to a shorter (but with more subjects) Repomse database showed that it was not good enough to use just a general model from CHB-MIT. Namely, seizure models from CHB-MIT showed low sensitivity (TPR) and F1 scores on the Repomse database. However, personalizing seizure models using Repomse data improved predictions significantly, leading to improved results compared to generalized (or even personalized) Repomse models. This means that for epilepsy detection on the Repomse dataset, it is possible to reuse non-seizure models from the CHB-MIT database and only retrain seizure models.
    When transferring knowledge from Repomse to CHB-MIT, an opposite effect was noticed. Seizures from Repomse were general enough, but non-seizure models were not general enough to use on a dataset that contained much more non-seizure data. Thus, to avoid full training on CHB-MIT, seizure models from Repomse can be used. 
    All these results show that having a big long-term database with many subjects, and building generalized models trained on that database could be potentially reusable for new unseen patients (without having to retrain models). If needed, as in \textit{NSgen-Spers} and \textit{NSpers-Sgen} hybrid approaches, only one class could be retrained, saving a lot of training time.

    \vspace{2mm}
    \textbf{Limitations and future work}
    \\
    The Repomse dataset is very interesting due to the high number of subjects available, which enabled us to study properly generalized models. As we have shown, we indeed needed more than 40 subjects to achieve stable generalized models. This is more than what is available in other public datasets (i.e., 24 subjects in CHB-MIT or 18 in SWEC-ETHZ~\cite{burrello_laelaps_2019}). However, it is a short-term database, meaning that performances might be overestimated with respect to what they would be on a long-term database. Thus, we studied knowledge transfer between slightly bigger data subset from the CHB-MIT database, using 10 times more non-seizure data. Ideally, in the future, a similar analysis should also be performed on a long-term database with a similarly large number of subjects. 

    From a neurological perspective, as the CHB-MIT database contains pediatric patients and the Repomse adult population, mixing the types of patients might not be that compelling. Furthermore, databases contain a broad range of seizure types but, unfortunately, no labels. Mixing different seizure types adds a certain level of complexity. Here we were interested in the performance possible even in this limited scenario, but in the future, repeating this analysis separately for different types of seizure would be more interesting from a medical perspective.

    \vspace{-2mm}
    \section{Conclusion} 
    \label{sec:Conclusion}
    \vspace{-2mm}
    In this work, we have demonstrated how HD computing, and the way its models are built and stored, can be used to further understand, compare, and create more complex machine learning models. These possibilities are not feasible with other state-of-the-art models, such as random forests or neural networks. 
    We compare inter-subject similarity of seizure and non-seizure epilepsy models and then study the process of creation of generalized models from personalized ones, which is an essential part of distributed wearable applications. We tested a novel hybrid approach to create models that consist partially of personalized and partially of generalized models. This is a unique possibility that hyperdimensional computing allows, and we show that these models result in an improved epilepsy detection performance. Similarly, we use hybrid models to test knowledge transfer between models of different epilepsy databases. Finally, all those examples could be very interesting not only from an engineering perspective to create better models for wearables but also from a neurological perspective to understand individual epilepsy patterns better. 
 



	\small
	\printbibliography
\end{document}